\documentclass[runningheads]{llncs}

\usepackage{eccv}

\usepackage{eccvabbrv}

\usepackage{graphicx}
\usepackage{booktabs}

\usepackage[accsupp]{axessibility}  

\usepackage[pagebackref,breaklinks,colorlinks,citecolor=eccvblue]{hyperref}

\usepackage{orcidlink}

\usepackage{boldline}
\usepackage{relsize}
\usepackage{makecell}
\usepackage{arydshln}
\usepackage{amssymb}
\usepackage{subcaption}
\usepackage{mathtools}
\usepackage{adjustbox}
\usepackage{float}
\restylefloat{table}
\usepackage{graphicx}
\usepackage{indentfirst}

\newcommand{\veryshortarrow}[1][5pt]{\mathrel{%
   \hbox{\rule[\dimexpr\fontdimen22\textfont2-.2pt\relax]{#1}{.4pt}}%
   \mkern-4mu\hbox{\usefont{U}{lasy}{m}{n}\symbol{41}}}}
   
\newcommand{\mymakecell}[2]{\makecell{\normalfont #1 \\  $\scriptscriptstyle \boldsymbol{\pm}$ \scriptsize #2}}

\newcommand{\mymakecellh}[2]{\makecell{\normalfont #1$\scriptscriptstyle \boldsymbol{\pm}$\scriptsize #2}}

\begin{document}

\title{Trust And Balance: Few Trusted Samples Pseudo-Labeling and Temperature Scaled Loss for Effective Source-Free Unsupervised Domain Adaptation} 

\titlerunning{Trust And Balance Adaptation}

\author{Andrea Maracani\inst{1, 2, 3}\orcidlink{0000-0002-6217-8731} \and
Lorenzo Rosasco\inst{2}\orcidlink{0000-0003-3098-383X} \and
Lorenzo Natale\inst{3}\orcidlink{0000-0002-8777-5233}}

\authorrunning{Maracani et al.}

\institute{Istituto Italiano di Tecnologia, Genoa, ITALY \and
University of Genoa, Genoa, ITALY \\
\and
\email{andreamaracani@gmail.com}}

\maketitle

\begin{abstract}
  Deep Neural Networks have significantly impacted many computer vision tasks. However, their effectiveness diminishes when test data distribution (target domain) deviates from the one of training data (source domain). In situations where target labels are unavailable and the access to the labeled source domain is restricted due to data privacy or memory constraints, Source-Free Unsupervised Domain Adaptation (SF-UDA) has emerged as a valuable tool. Recognizing the key role of SF-UDA under these constraints, we introduce a novel approach marked by two key contributions: Few Trusted Samples Pseudo-labeling (FTSP) and Temperature Scaled Adaptive Loss (TSAL). FTSP employs a limited subset of trusted samples from the target data to construct a classifier to infer pseudo-labels for the entire domain, showing simplicity and improved accuracy. Simultaneously, TSAL, designed with a unique dual temperature scheduling, adeptly balance diversity, discriminability, and the incorporation of pseudo-labels in the unsupervised adaptation objective. Our methodology, that we name Trust And Balance (TAB) adaptation, is rigorously evaluated on standard datasets like Office31 and Office-Home, and on less common benchmarks such as ImageCLEF-DA and Adaptiope, employing both ResNet50 and ViT-Large architectures. Our results compare favorably with, and in most cases surpass, contemporary state-of-the-art techniques, underscoring the effectiveness of our methodology in the SF-UDA landscape.
\end{abstract}
\section{Introduction}
\label{sec:intro}

\begin{figure}[t]
    \centering
    \includegraphics[width=0.67\linewidth]{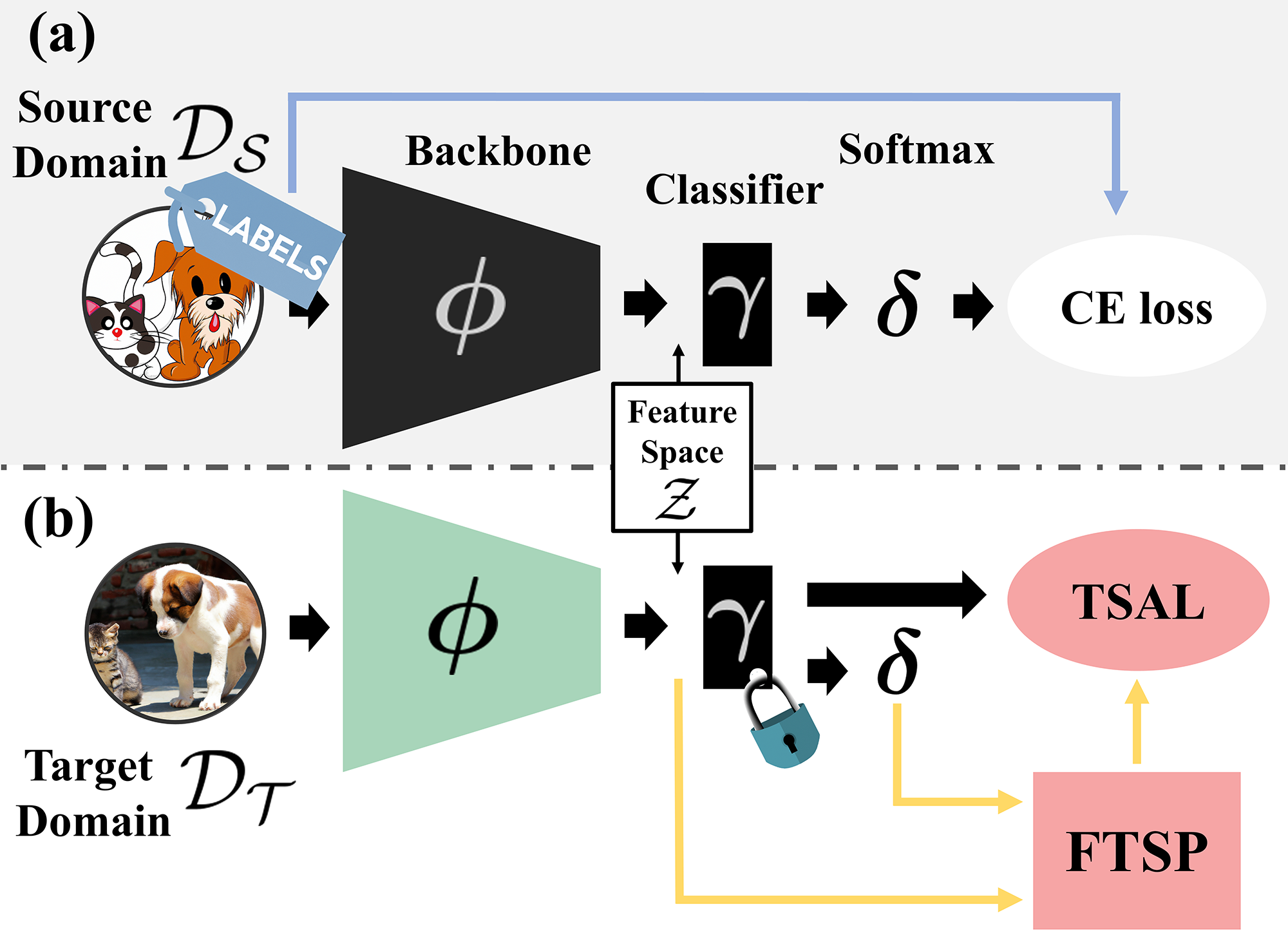}
    \caption{\textbf{SF-UDA Pipeline (our contributions in red).} In the upper section \textbf{(a)}, the source model is trained on the source domain through a conventional supervised method (indicated by the blue arrow). In the lower section \textbf{(b)}, adaptation to the target domain is conducted using our proposed pseudo-labeling method (FTSP) and objective function (TSAL), as shown by the yellow arrows. Consistent with the method of \cite{liang2020we}, the classifier $\gamma$ remains unchanged during the adaptation phase, while the backbone (in green) is adapted.}
    \label{fig:pipeline}
\end{figure}

Deep neural networks (DNNs) have made significant advancements in computer vision tasks, including image classification, detection, and semantic segmentation \cite{chai2021deep}. However, they often face challenges when the distribution of the test data, or the \textit{target domain}, differs from the training data, known as the \textit{source domain}. Such domain discrepancies, stemming from environmental changes, device variations or different image styles, limit the effectiveness of DNNs in real-world applications. \\
Unsupervised Domain Adaptation (UDA) aims to apply knowledge from a labeled source domain to an unlabeled target domain \cite{wilson2020survey}. While conventional UDA strategies demand access to both domains to mitigate the domain shift, there exist scenarios, especially in sensitive sectors like healthcare, where accessing the source data is constrained due to privacy or storage issues. This led to the advent of Source-Free Unsupervised Domain Adaptation (SF-UDA) in image classification \cite{liang2020we}, building upon ideas from Hypothesis Transfer Learning \cite{kuzborskij2013stability}. Essentially, SF-UDA leverages a model trained on the source, without a direct access to source data. Contemporary advances in SF-UDA encompass methodologies like entropy-minimization, generative modeling, class prototyping, self-training and many others \cite{fang2022source}. As we will see in Sec.~\ref{sec:related_work}, our approach shares some parallels with pseudo-label denoising and entropy-minimization techniques. 

In particular, we present a novel pseudo-labeling paradigm, \textbf{Few Trusted Samples Pseudo-labeling (FTSP)}, which accentuates simplicity and the quality of pseudo-labels. Unlike conventional, more complex, pseudo-labeling techniques, our method centers on creating a training set using a restricted subset of \textit{trusted} samples (i.e. with high likelihood to be correctly labeled by the source classifier) from the target domain (limited up to 3 samples per class). While our framework is agnostic to the choice of classifier, for simplicity, we adopted Multinomial Logistic Regression (MLR) in our main experiments and we present an ablation study with different classifiers in the supplementary material. Despite potential overfitting concerns with MLR on this limited dataset, it empirically demonstrates proficient generalization capabilities across the broader target domain, effectively inferring high-quality pseudo-labels. We also propose a pseudo-label refinement phase, including a deletion mechanism based on classifier uncertainty and a pseudo-label completion step via \textit{Label Spreading} \cite{zhou2003learning}.

The analysis of Yang et al.~\cite{yang2022attracting} emphasized that most SF-UDA methods revolve around an objective involving two core components: a \textit{diversity term} for prediction variability and a \textit{discriminability term} to enhance target samples differentiation. Inspired by Information Maximization objective of SHOT~\cite{liang2020we} we propose the \textbf{Temperature Scaled Adaptive Loss (TSAL)}: a novel and advanced objective to guide the adaptation process. In particular TSAL is specifically designed to use a dual temperature scheduling to dinamically balance the discriminability, diversity and the incorporation of pseudo-labels and their significance throughout the whole adaptation phase, showing improved performance in SF-UDA. In summary, our key contributions are:

\begin{itemize}
\item \textbf{Few-Trusted Samples Pseudo-labeling (FTSP)}: an effective pseudo-labeling technique involving the training of a classifier employing a curated very-limited subset of \textit{trusted samples} from the target domain. We further propose incorporating pseudo-label deletion and completion steps (with Label Spreading) for additional refinement.

\item \textbf{Temperature Scaled Adaptive Loss (TSAL)}: our advanced balance strategy to effectively calibrate the equilibrium between diversity, discriminability, and pseudo-label significance in the objective, resulting in enhanced SF-UDA results.

\item \textbf{Robust Benchmarking and Analysis}: our method undergoes rigorous evaluations on standard datasets like Office31 and Office-Home, and on emerging benchmarks such as ImageCLEF-DA and Adaptiope, using both ResNet50 and ViT-Large. Beyond traditional single-seed evaluations, we present a multi-seed robustness analysis (5 seeds) and recreate some selected state-of-the-art techniques for a thorough comparative insight.

\end{itemize}

\noindent The structure of this paper is as follows: \cref{sec:related_work} provides an overview of pertinent literature. The SF-UDA setting is presented in \cref{sec:problem}. The proposed methodology is delineated in \cref{sec:method}. Experimental procedures and results are detailed in \cref{sec:experiments}. Concluding remarks are presented in \cref{sec:conclusion}. Additional details and ablation studies are presented in the supplementary material, while the code will be released at \url{https://github.com/andreamaracani/TAB_SFUDA} 


\section{Related Work}
\label{sec:related_work}

\indent\textbf{Unsupervised Domain Adaptation (UDA).} UDA aims to adapt models from a source domain (with available labels) to an unlabeled target domain. The foundational principles of UDA are rooted in the theoretical works by Mansour et al.~\cite{mansour2009domain} and Ben-David et al.~\cite{ben2010theory}. Early methods include sample selection~\cite{huang2006correcting} and feature projection~\cite{pan2010domain}, followed by techniques designed to adapt Deep Neural Networks, such as adversarial training~\cite{ganin2016domain}, Maximum Mean Discrepancy~\cite{kang2019contrastive}, Bi-directional Matching~\cite{na2021fixbi}, Margin Disparity Discrepancy~\cite{zhang2019bridging} and many others~\cite{wilson2020survey}. Though initially centered on image classification, UDA has expanded to include tasks like object detection~\cite{oza2023unsupervised} and semantic segmentation~\cite{toldo2020unsupervised}. A notable challenge in UDA is the need for simultaneous access to both source and target data during training, which may be a nuisance or even impracticable in some contexts, e.g. due to intellectual property or privacy issues. \\
\indent\textbf{Source-Free UDA (SF-UDA).} As a subdomain of UDA, SF-UDA negates the direct access to source domain data during adaptation. The field gained traction following Liang et al.~\cite{liang2020we}. Thereafter, a multitude of methods emerged, achieving interesting results on common UDA benchmarks. Noteworthy SF-UDA techniques include generative model-driven methods like 3C-GAN~\cite{li2020model}, algorithms based on the feature space's neighborhood structure (e.g., NRC~\cite{yang2021exploiting} and AAD~\cite{yang2022attracting}), methods transferring Batch Normalization statistics~\cite{yang2022source, hou2020source}, strategies constructing surrogate source domains during adaptation~\cite{tian2021source,ding2023proxymix}, techniques utilizing knowledge distillation within a mean-teacher~\cite{tarvainen2017mean} paradigm~\cite{liu2021graph, liu2022source, yang2021transformer}, and those incorporating Contrastive learning~\cite{liu2022source, agarwal2022unsupervised, zhao2022adaptive}. A comprehensive review of contemporary SF-UDA approaches can be found in~\cite{fang2022source}. \\
\indent\textbf{Learning with pseudo-labels.} SF-UDA methods often necessitates the creation of target pseudo-labels for improved training. However, the potential presence of errors in these pseudo-labels parallels training with noisy labels. Numerous methods aim to contrast the potential noise-fitting caused by these inaccuracies. Notable approaches encompass the utilization of reliable labels through co-teaching dual networks~\cite{han2018co}, Negative Learning (NL) implementation~\cite{kim2019nlnl}, and the adoption of noise-resistant loss functions~\cite{englesson2021generalized}. In the SF-UDA setting, Zhang et al.~\cite{zhang2021unsupervised} advanced a technique that refines noise rate estimation and emphasizes early-stage sample retention. Luo et al.~\cite{luo2021exploiting} presented a method to rectify pseudo-label errors using negative learning, tailored for semantic segmentation. Yang et al.~\cite{yang2022divide} fused pseudo-label denoising with self-supervised knowledge distillation.  
\noindent Litrico et al.~\cite{litrico2023guiding} integrated insights from nearest neighbors and entropy-based uncertainty estimation, further augmented by a temporal queue mechanism and self-learning methodologies. \\
\indent Related to our work, there are also some methods that utilize significant samples for the adaptation~\cite{tian2023source, yang2023auto, xu2019unsupervised}. However, our pseudo-labeling strategy, detailed in Sec.~\ref{sub:ftsp}, distinctly diverges from these approaches, offering a unique methodological contribution. Additionally, while many contemporary techniques lean toward complexity, our methodology distinguishes itself through its efficiency and effectiveness, surpassing in performance also more complex techniques.
Even if we might optionally utilize the well-established Label Spreading to alleviate label noise, the essence of our method lies in generating inherently accurate pseudo-labels with a classifier trained on a meticulously selected set of a very limited number of trusted target samples. Additionally, the harmonious integration of discriminability and diversity in our TSAL objective further enhances the method's robustness against pseudo-label noise. As detailed in \cref{sec:experiments}, our approach consistently aligns with or even surpasses state-of-the-art (SOTA) performance across benchmarks, asserting the robustness of our loss function to pseudo-label noise.

\section{Problem Definition} 
\label{sec:problem}

Before presenting our proposed approach, we set the foundation for SF-UDA in the context of image classification. Let \(\mathcal{X} \in \mathbb{R}^{H \times W \times 3}\) denote the space of RGB images with height \(H\) and width \(W\). The label space, covering \(C\) distinct categories, is represented by \(\mathcal{Y} = \{c\}_{c=1}^{C}\). We postulate two distinct distributions over \(\mathcal{X} \times \mathcal{Y}\): the source domain \(\mathcal{D}_S\) and the target domain \(\mathcal{D}_T\). We consider the Close-set assumption: the label space remains consistent between these domains, guaranteeing that each category possesses a non-zero probability of manifestation in both.

Consider \(f_{\boldsymbol{\theta}}: \mathcal{X} \to \mathcal{Y}\), a function parametrized by \(\boldsymbol{\theta}\), which maps each input image to its associated label in \(\mathcal{Y}\). The main objective in both UDA and SF-UDA is to identify this function along with its optimal parameters, ensuring accurate target domain predictions. While Deep Neural Networks are the prevalent choice for this function, data restrictions depend on the specific adaptation setting. Specifically, the SF-UDA framework consists of \textbf{two stages} (see Fig.~\ref{fig:pipeline}): 
\begin{enumerate}
    \item A labeled dataset from the source distribution, \(\mathcal{S}=\{(\mathbf{x}_S^{(i)}, y_S^{(i)})\}_{i=1}^{M} \sim \mathcal{D}_S^M\), is employed to determine the function parameters \(\boldsymbol{\theta}_S\) such that the function performs optimally on the source domain.

    \item The source dataset becomes inaccessible, though the parameters \(\boldsymbol{\theta}_S\) remain available together with an unlabeled dataset from the target domain (marginal) distribution, represented as \(\mathcal{T}=\{\mathbf{x}_T^{(i)}\}_{i=1}^{N} \sim \mathcal{D}_T^N(\mathcal{X})\). This is employed to adjust the model parameters to \({\boldsymbol{\theta}}_T\), with the goal of obtaining an improved performance on the target domain.

\end{enumerate}

\subsection{Architecture}

In alignment with the conventions established in earlier studies, the function \(f\) (we omit parameters $\boldsymbol{\theta}$ for notation simplicity) is articulated as an composition of multiple functions, as illustrated in Fig.~\ref{fig:pipeline}:

\begin{equation}
    f(\mathbf{x}) \mapsto \arg\max_{c \in \mathcal{Y}}\{\delta(\gamma(\phi(\mathbf{x})))_c\} = \hat{y}
\end{equation}

 where function \(\phi: \mathcal{X} \to \mathcal{Z} \subset \mathbb{R}^{d}\) is the backbone and it operates as a feature extractor mapping images into the $d$-dimensional feature space $\mathcal{Z}$. $\gamma: \mathcal{Z} \to \mathbb{R}^C$ is as a classifier that maps features into the $C$-dimensional space of logits. Lastly, $\delta: \mathbb{R}^C \to \Delta^{C-1}$ denotes the Softmax function, which translates logits into the $C-1$ simplex that signifies classification probabilities for each class. The ultimate prediction class $\hat{y}$ is extracted using the $\arg\max$ operation on these probability values.
\section{Method}

\label{sec:method}
A fundamental guiding principle in our method design is ensuring backbone independence. While specialized architectural modifications, such as adapting Batch Normalization layers, freezing some specific layers, or introducing specific additional modules, can offer advantages in certain benchmarks (e.g., with ResNet50), we deliberately avoid them. This decision is rooted in our understanding that in scenarios extending beyond typical benchmarks, more advanced models could be employed within the SF-UDA framework. Therefore, our ambition is to devise a universally applicable solution.
We present an overview of our proposed method and in the next sections we will give a detailed description of the algorithm.

\noindent \textbf{Stage 1: source fine-tuning.} We initiate training using a pre-trained (e.g., on ImageNet) feature extractor and we adopt an end-to-end fine-tuning approach, adjusting the backbone's weights with the labeled source dataset in alignment with previous SF-UDA algorithms, leveraging insights from \cite{maracani2023key} that highlight the benefits of such source fine-tuning.

\noindent \textbf{Stage 2: target adaptation.} When unlabeled target data becomes available the model undergoes unsupervised self-training.
At the beginning of each epoch, pseudo-labels for the entire target domain are reassessed using our FTSP methodology (Sec.~\ref{sub:ftsp}), then our TSAL objective is minimized to balance diversity, discriminability and pseudo-labels significance with a dual temperature scaling (Sec.~\ref{sub:TSAL}). During this adaptation phase, the weights of the backbone $\phi(\cdot)$ are updated while the classifier, $\gamma(\cdot)$, remains unchanged in consistency with \cite{liang2020we}.

\vspace{-13pt}
\begin{figure}[h!]
    \centering
    \includegraphics[width=0.99\linewidth]{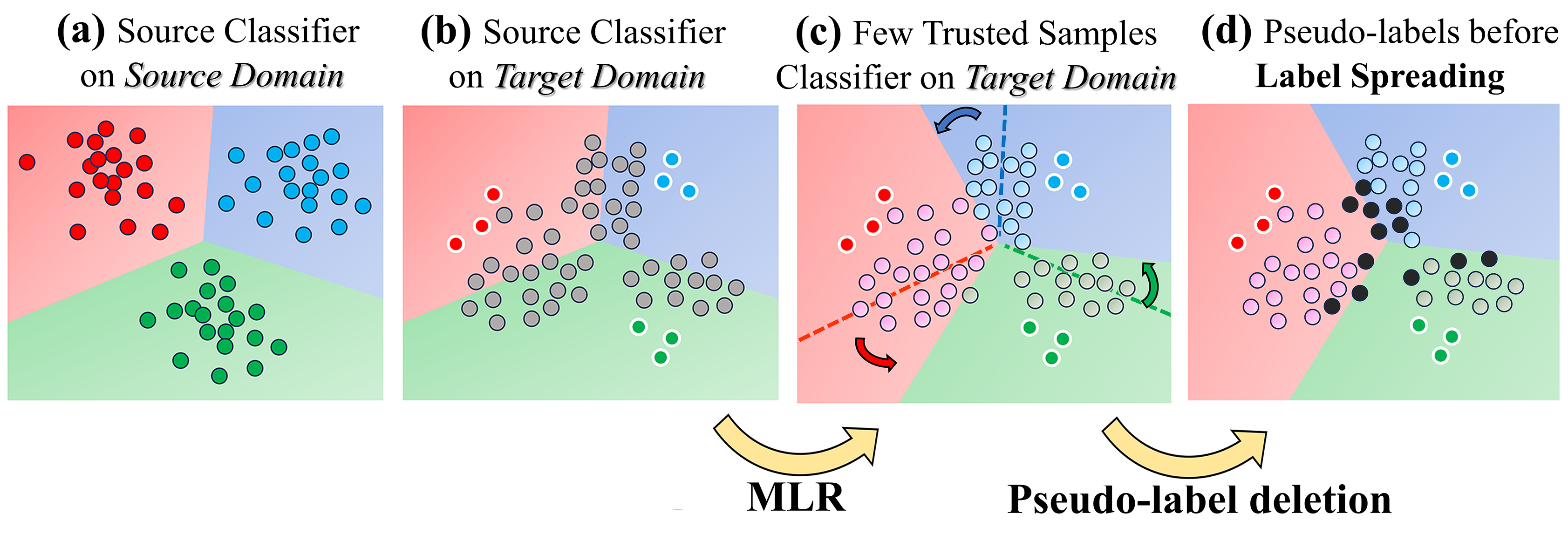}
    \caption{\textbf{Pseudo-labeling with Few Trusted Samples}:
\textbf{(a)} Classifier trained on the source domain demonstrates robust performance within the same domain. \textbf{(b)} The same classifier underperforms on the unlabeled target domain (represented by grey dots). A minimal set of trusted target samples (indicated by colored dots with white outlines) is selected, being deemed most likely to be correctly classified. \textbf{(c)} Using these few trusted samples, a Multinomial Logistic Regression (MLR) classifier is trained, leading to decision boundaries that align more closely with the target domain and subsequently providing pseudo-labels for the entire target domain. \textbf{(d)} A fraction of uncertain pseudo-labels is eliminated prior to the application of Label Spreading, finalizing the Few Trusted Samples Pseudo-labeling (FTSP) process.}
        \label{fig:ftsp}
\end{figure}

\subsection{Pseudo-labeling through few trusted samples}\label{sub:ftsp}

Our algorithm's development was heavily influenced by a clear insight: the selection of an extremely limited number of high-quality target domain samples can lay a foundation for constructing a classifier that surpasses the performance of the original source classifier $\gamma$. This perspective deviates from traditional methods that often rely on large sample sizes or intricate techniques. By identifying a restricted set of $K$ \textbf{trusted samples} (TS) for each class (i.e. samples that are very likely to be correctly classified), we build a classifier using the combined dataset with $K \times C$ samples, providing a strong basis for predictions across the target domain.

\textbf{Trusted samples training set.} In our quest for a simplified methodology, for each class $c \in \mathcal{Y}$, we choose the $K$ feature samples with the top predicted probabilities according to the source classifier \(\delta(\gamma(\cdot))\):

\begin{equation}
    \text{TS}_c \coloneqq \{\textbf{z}^{(1)}_c, \ldots, \textbf{z}^{(K)}_c\} = argmax^K_{\textbf{z} \in \mathcal{Z}_T}\{\delta(\gamma(\textbf{z}))_c\}
\end{equation}

To clarify, the notation $argmax^K$ denotes the function returning the $K$ arguments with the greatest values. $\mathcal{Z}_T$ represents the set of all target features (evaluated with backbone $\phi$), and $\delta(\gamma(\mathbf{z}))_c$ indicates the predicted probability of the feature $\mathbf{z}$ being categorized into class $c$ by the source classifier. Repeating this for each class produces a \textit{few-trusted-samples} training set with known labels (that are likely to be correct).

\textbf{Trusted samples classifier.} Considering this dataset, we first normalize the features vectors and then we train a simple classifier from scratch, subsequently deploying it to infer pseudo-labels for the entire target domain. Our method is independent of the chosen classifier; however, in our experiments, we consistently employed by default Multinomial Logistic Regression (MLR). While the results highlight the efficacy of MLR (as elaborated in Sec.~\ref{sub:results}), we acknowledge its potential limitations. To further explore these aspects, a post hoc ablation study is provided in the supplementary material considering different classifiers and hyperparameters. This study indicates that Linear Discriminant Analysis (LDA) may offer additional advantages to our approach.

\textbf{Pseudo-label refinement.} For improved pseudo-label quality, we propose an additional refinement phase in our algorithm. Specifically, we introduce a pseudo-label deletion step, which entails removing a certain percentage (per class) of the least certain pseudo-labels, based on the MLR classifier output probabilities. This is followed by a pseudo-label completion step using the established semi-supervised learning method of Label Spreading \cite{zhou2003learning}. These procedures are useful to reassess and enhance the overall label consistency. Their advantages are explored in the ablation study in Sec.~\ref{sub:analysis} and in the supplementary material. 

We refer to our distinctive pseudo-labeling technique as \textbf{Few Trusted Samples Pseudo-labeling (FTSP)}, illustrated in Fig.~\ref{fig:ftsp}. 

\begin{figure}
\centering
\begin{subfigure}[t]{0.6\columnwidth}
    \includegraphics[width=\linewidth]{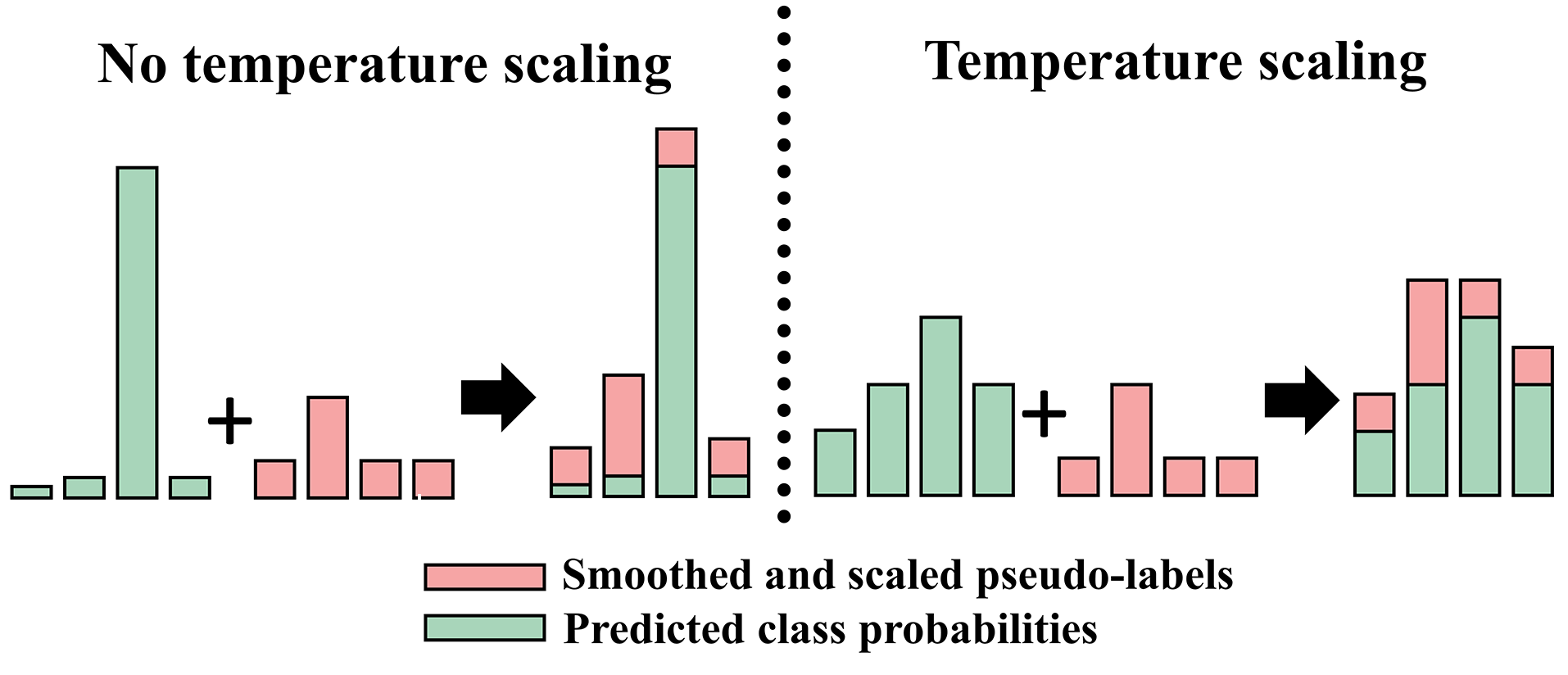}
    \caption{Temperature scaling in the discrimability term.}
    \label{subfig:temperature_pseudo}
\end{subfigure}
\hspace{0.1cm}
\begin{subfigure}[t]{0.3\columnwidth}
    \includegraphics[width=\linewidth]{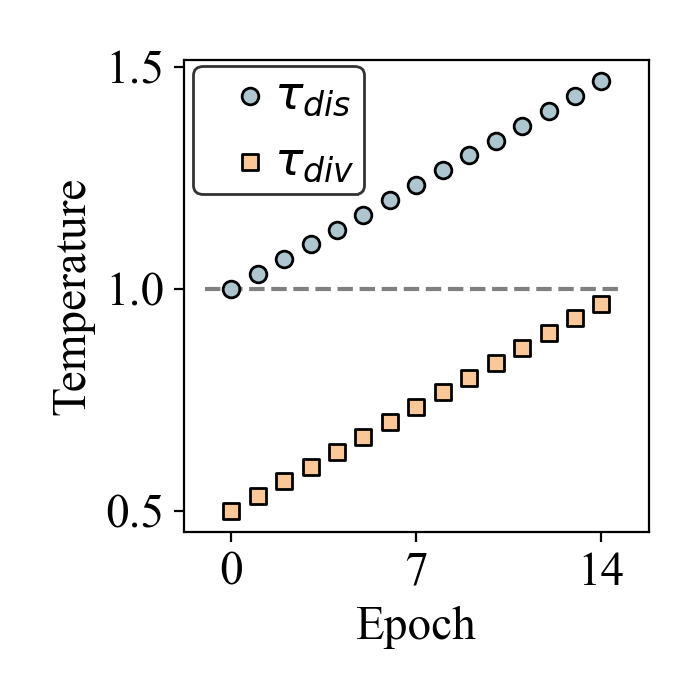}
    \caption{Temperature schedules.}
    \label{subfig:temperature_schedule}
\end{subfigure}
  \caption{\textbf{(a)} the temperature scaling in our discrimability term enables a fair competition between model predictions and the pseudo-labels at the end of the training when the network becomes overconfident. \textbf{(b)} our schedules $\tau_{\text{dis}}(\cdot)$ and $\tau_{\text{div}}(\cdot)$ respectively for the discrimanbility and diversity term.}
\end{figure}

\subsection{Temperature scaled loss for adaptive training}\label{sub:TSAL}

\textbf{Intuition and motivation.} The analysis in \cite{yang2022attracting} shows that most SF-UDA objectives can be delineated into two primary goals. The first is to enhance prediction distinction (discriminability term, \textit{dis}), and the second is to diversify these predictions (diversity term, \textit{div}). 

\begin{equation}
    loss_\text{SF-UDA} = dis + div
\end{equation}

In particular, the Information Maximization (IM) objective of SHOT \cite{liang2020we} has exhibited consistent performance across diverse architectures and datasets \cite{maracani2023key}. Such robustness is not universally observed among all state-of-the-art UDA and SF-UDA methods, as highlighted in Kim et al.'s study \cite{kim2022broad}. Nevertheless, we have observed some limitations and weaknesses of the SHOT objective:

\begin{itemize}
    \item The discriminability component uses both the model's current predictions (to minimize the entropy) and some pseudo-labels pre-computed through clustering. But as training moves forward, the model becomes more sure of its own predictions. This increased (over-)confidence can make it harder to adjust predictions based on pseudo-labels (see Fig.~\ref{subfig:temperature_pseudo}).
    \item The diversity term is represented by the negative entropy of the average output probabilities. Early in the adaptation process, under common domain shifts, the network often lacks confidence in its predictions, resulting in an already high average entropy (so a very low negative entropy). This can cause the discriminability term to be overly emphasized in the initial stages.
\end{itemize}

To address these challenges we design a new objective that incorporates a dual-temperature scaling approach to balance discriminability, diversitity and pseudo-label significance across the whole adaptation process. At the start, we use a standard temperature value ($=1$) for the discriminability term. As the model becomes more confident, we increase the temperature ($>1$) to moderate the model's growing certainty.
Conversely, for the diversity term, we adopt a lower temperature ($<1$) to refine predictions early in training, transitioning to a standard temperature ($=1$) towards the training's conclusion. We now present the designed objective encapsulating these insights.

\textbf{Temperature Scaled objective.} For a batch comprising $B$ images, denoted as $\mathcal{B} = \{\mathbf{x}^{(i)}\}_{i=1}^B$, the model discerns the output logit vectors $\hat{\mathcal{L}} = \{\mathbf{\hat{l}}^{(i)}\}_{i=1}^B$. Further, one-hot pseudo-labels are computed through our FTSP to yield $\hat{\mathcal{Y}} = \{\mathbf{\hat{y}}^{(i)}\}_{i=1}^B$. An initial step entails the softening of pseudo-labels to mitigate erroneous pseudo-label impacts, resulting in the smooth pseudo-label set $\hat{\mathcal{Y}}_S$. Specifically, we utilize conventional label smoothing with a default factor $S$ of $0.1$:

\begin{equation}
    \mathbf{\hat{y}}^{(i)}_S = \mathbf{\hat{y}}^{(i)} \cdot (1-S) + \mathbf{1}_C \cdot S/C
\end{equation}

where $\mathbf{1}_C$ is a $C$-dimensional vector containing $1$s.
We now construct an objective target distribution for a generic target sample $\mathbf{x}_i$ as a mixture of temperature scaled predicted probability and the pseudo-label:

\begin{equation}
    \mathbf{\hat{q}}^{(i)}(t) = \delta\left( \frac{ \mathbf{\hat{l}}^{(i)}}{\tau_{\text{dis}}(t)} \right) + \alpha \cdot  \mathbf{\hat{y}}_S^{(i)} 
\end{equation}

Where $\alpha$ is a constant set to $0.3$ and $\tau_{\text{dis}}(\cdot)$ is the temperature function in order to scale the predicted probabilities and make them softer at the end of the training. The $t$ variable, in our schedule (that we will discuss shortly) is an integer corresponding to the number of epoch. The integration of both the predictions of the network (self-regularization) and the pseudo-labels in the objective distribution enables the competition between model predictions and pseudo-labels computed with FTSP. The loss's \textbf{discriminability term} is hence:

\begin{equation}
    dis(\hat{\mathcal{L}}, \hat{\mathcal{Y}_S};t) \coloneqq \frac{1}{B} \sum_{i=1}^B H(\mathbf{\hat{q}}^{(i)}(t), \delta(\mathbf{\hat{l}}^{(i)}))
\end{equation}
where $H(\cdot, \cdot)$ is the cross-entropy. 
For diversity, a temperature scaled variant of \cite{liang2020we} is employed. Let define the output average (scaled) probability as:

\begin{equation}
\mathbf{\bar{p}}(t) \coloneqq \frac{1}{B} \sum_{i=1}^B \delta\left( \frac{\mathbf{\hat{l}}^{(i)}}{\tau_{\text{div}}(t)} \right)
\end{equation}

Where $\tau_{\text{div}}(\cdot)$ is the second temperature schedule function in order to scale the predicted probabilities and make them sharpen at the beginning of the adaptation procedure. Then the \textbf{diversity term} is:

\begin{equation}
    div(\hat{\mathcal{L}}; t) \coloneqq -H(\mathbf{\bar{p}}(t))
\end{equation}
where $H(\cdot)$ is the entropy function. 

The overall objective, that we refer to as \textbf{Temperature Scaled Adaptive Loss (TSAL)} is: 

\begin{equation}
    loss_{\text{TSAL}}(\hat{\mathcal{L}}, \hat{\mathcal{Y}_S}; t) \coloneqq dis(\hat{\mathcal{L}}, \hat{\mathcal{Y}_S}; t) + div(\hat{\mathcal{L}}; t)
\end{equation}

\noindent\textbf{Temperature Scaling schedule.} As shown in Fig.~\ref{subfig:temperature_schedule} the functions $\tau_{\text{dis}}(\cdot)$ and $\tau_{\text{div}}(\cdot)$ undergo adjustments every epoch. While $\tau_{\text{dis}}(\cdot)$ gradually enhances prediction softness, $\tau_{\text{div}}(\cdot)$ initially sharpens predictions, only to soften them towards the training's closure. The essence of these functions is rooted in the preliminary motivations. Specifically, $\tau_{\text{dis}}(\cdot)$ transitions linearly from 1 to 1.5, whereas $\tau_{\text{div}}(\cdot)$ moves from 0.5 to 1 (with 1 signifying no temperature modulation).

\section{Experimental results}
\label{sec:experiments}
We evaluate the proposed approach, that we name \textbf{Trust And Balance (TAB)}, and compare it with SOTA methods for SF-UDA on image classification. 

\subsection{Setup}\label{sub:setup}

\indent\textbf{Datasets.} For our evaluation, we chose a combination of widely-recognized datasets (Office31 and Office-Home), as well as datasets that are slightly less prevalent in typical benchmarks (Adaptiope and ImageCLEF-DA). This selection underscores the versatility of our method. \textit{Office-31}~\cite{saenko2010adapting}: this dataset features $4\,110$ images and includes three domains: Amazon (A), DSLR (D), and Webcam (W). \textit{Office-Home}~\cite{venkateswara2017deep}: a medium-scale dataset that comprises $15\,500$ images, partitioned into 65 categories and spread across 4 domains: Art (A), Clip Art (C), Product (P), and Real World (R). \textit{Adaptiope}~\cite{ringwald2021adaptiope}: a large-scale dataset containing $36\,900$ images. It is categorized into 123 classes and spans three domains: Product (P), Real Life (R), and Synthetic (S). \textit{ImageCLEF-DA}~\cite{long2017deep}: a small dataset including $2\,400$ images. It is divided into 12 classes and 4 domains: Bing (B), Caltech (C), ImageNet (I), and Pascal (P).

\indent\textbf{Backbones.} For a fair comparison with a wide range of other popular SOTA methods we adopted ResNet50 \cite{he2016deep} (pre-trained on ImageNet~\cite{deng2009imagenet}) for our experiments and the typical single-run results. To evaluate the robustness of our approach we further investigate multi-run results (we use 5 seeds in the robustness analysis) and we adopted also a better performing architecture to prove the versatility of our approach, namely ViT-Large \cite{dosovitskiy2020image} (pre-trained on ImageNet21k). To have a comparison we selected 3 popular SOTA SF-UDA methods that we recognize as easily reproducible and we run them through our robustness analysis: SHOT~\cite{liang2020we}, AAD~\cite{yang2022attracting} and NRC~\cite{yang2021exploiting}. 

\indent\textbf{Implementation Details}. Our method is developed using the PyTorch~\cite{paszke2019pytorch} framework and adheres to the standard guidelines and hyperparameters found in the SF-UDA literature, such as \cite{liang2020we}, \cite{yang2021exploiting}, and \cite{yang2022attracting}. We employ the SGD optimizer for training, configured with a momentum of $0.9$, weight decay of $10^{-3}$, batch size of $64$, and an input image dimension of $224 \times 224$. The pre-trained backbone is enriched by a newly initialized bottleneck layer that maps features to 256 dimensions and the final classifier. The initial learning rates are set to $10^{-3}$ for the backbone and ten times higher for both the bottleneck and classifier. These rates then follow exponential scheduling throughout training. Notably, the classifier's weights are frozen during the adaptation phase. Additionally, we incorporate MixUp regularization~\cite{zhang2017mixup} throughout the training process. For FTSP we use a value of $K=3$ for ResNet50 and $K=7$ for ViT-L (accounting for the more precise predictions of this advanced architecture) and a Multinomial Regression Classifier. In the label deletion step we delete, for each class, the $20\%$ of less confident pseudo-labels, and then we apply Label Spreading. Depending on the computational needs of various experiments, we utilized either Nvidia V100 16GB or Nvidia A100 80GB GPUs. For comprehensive details, including an analysis of our approach's efficiency and a detailed breakdown of its \textbf{computational requirements}, please refer to the supplementary material.

\vspace{-0.5cm}
\begin{table}[h!]
  \centering
  \setlength{\tabcolsep}{5pt}

\begin{adjustbox}{width = 0.7\linewidth, center}
\begin{tabular}{lccccccc}
    \toprule
    \textbf{Method} & A$\veryshortarrow$D &  A$\veryshortarrow$W & D$\veryshortarrow$A &  D$\veryshortarrow$W & W$\veryshortarrow$A &  W$\veryshortarrow$D & \textbf{Avg} \\
    \midrule
    ResNet50 \cite{he2016deep}& 68.9 & 68.4 & 62.5 & 96.7 & 60.7 & 99.3 & 76.1 \\
    3C-GAN$_{\text{R50}}$~\cite{li2020model} & 92.7 & 93.7 & 75.3 & 98.5 & \textbf{77.8} & 99.8 &  89.6 \\
    BNM-S$_{\text{R50}}$ \cite{cui2021fast} & 93.0 & 92.9 & 75.4 & 98.2 & 75.0 & \underline{99.9} & 89.1 \\
    SHOT$_{\text{R50}}$ \cite{liang2020we} & 94.0 & 90.1 & 74.7 & 98.4 & 74.3 &  \underline{99.9} &  88.6 \\
    AAD$_{\text{R50}}$ \cite{yang2022attracting}& \underline{96.4} & 92.1 & 75.0 &  \underline{99.1} & 76.5 &  \textbf{100.0} & \underline{89.9} \\
    NRC$_{\text{R50}}$ \cite{yang2021exploiting} & 96.0 & 90.8 & 75.3 & 99.0 &  75.0 & \textbf{100.0} &  89.4 \\
    
    DIPE$_{\text{R50}}$~\cite{wang2022exploring} & \textbf{96.6} & 93.1 & 75.5 & 98.4 & \underline{77.2} & 99.6 & \textbf{90.1} \\

    A$^2$Net$_{\text{R50}}$~\cite{xia2021adaptive} & 94.5 & \underline{94.0} & \underline{76.7} & \textbf{99.2} & 76.1 & \textbf{100.0} & \textbf{90.1} \\
    \hdashline\noalign{\vskip 0.5ex}
    \textbf{TAB}$_{\text{R50}}$ & 94.4 & \textbf{94.7} & \textbf{76.9} & 97.4 & 76.0 & 99.8 & \underline{89.9} \\
    \midrule
    ViT-L \cite{dosovitskiy2020image} & 91.8 & 94.1 & 80.5 & 98.5 & 86.7 & \underline{99.6} & 91.4 \\
    SHOT$_{\text{ViT}}$ \cite{liang2020we} & 98.2 & 97.9 & 82.9 & 97.2 & 85.7 &  \textbf{99.8} &  93.6 \\
    AAD$_{\text{ViT}}$ \cite{yang2022attracting}& \underline{98.8} & \underline{98.5} & 79.8 & \underline{99.3} & 84.4 & \textbf{99.8} & 93.4 \\
    NRC$_{\text{ViT}}$ \cite{yang2021exploiting} & 98.0 & 98.1 & \underline{85.9} & 99.0 & \textbf{87.0} & \textbf{99.8} & \underline{94.6} \\
    \hdashline\noalign{\vskip 0.5ex}
    \textbf{TAB}$_{\text{ViT}}$  & \textbf{100.0} & \textbf{98.9} & \textbf{86.4} & \textbf{99.9} & \underline{86.9} & \textbf{99.8} & \textbf{95.3} \\
    \bottomrule
\end{tabular}
\end{adjustbox}
\caption{Comparison of SOTA methods on the \textit{Office31} dataset using ResNet50 and ViT-L backbones. Each column represents an experiment SRC$\veryshortarrow$TGT, while the rightmost column provides the average accuracy. The top results are highlighted in \textbf{bold}, while the runners-up are \underline{underlined}. All ViT-L outcomes were independently obtained by us. \textbf{Note:} results for TAB are presented without any dataset-specific selection of hyperparameters in order to offer a valuable assessment. As detailed in the supp. material's ablation study, TAB can achieve a 90.3\% accuracy on Office-31 with ResNet50.}
  \label{tab:office31}
\end{table}

\vspace{-1cm}
\subsection{Results}\label{sub:results}

\indent\textbf{Office31.} The results for the Office31 benchmark are presented in Table~\ref{tab:office31}. Our approach yields results that are competitive with SOTA methods when using the ResNet50 architecture. Additionally, when employing the advanced ViT-L architecture, our method surpasses the performance of the considered techniques, achieving an average accuracy of 95.3\%.

\indent\textbf{Office-Home.} As detailed in Table~\ref{tab:officehome}, our method's outcomes on the Office-Home benchmark are either on par or superior to SOTA methods using ResNet50. Moreover, with the ViT-L architecture, our method outperforms other techniques, achieving an average accuracy of 88.2\%.

\begin{table*}[h!]
  \centering
  \setlength{\tabcolsep}{2pt}

  \begin{adjustbox}{width = 0.99\textwidth, center}
  \begin{tabular}{lccccccccccccc}
    \toprule
    \textbf{Method} & A$\veryshortarrow$C &  A$\veryshortarrow$P & A$\veryshortarrow$R &  C$\veryshortarrow$A &  C$\veryshortarrow$P & C$\veryshortarrow$R &  P$\veryshortarrow$A &  P$\veryshortarrow$C & P$\veryshortarrow$R &  R$\veryshortarrow$A & R$\veryshortarrow$C & R$\veryshortarrow$P & \textbf{Avg} \\
    \midrule
    ResNet50 \cite{he2016deep} & 46.3 & 67.5 & 75.9 & 59.1 & 59.9 & 62.7 & 58.2 & 41.8 & 74.9 & 67.4 & 48.2 & 74.2 & 61.3\\
    G-SFDA$_{\text{R50}}$~\cite{yang2021generalized} & 57.9 & 78.6 & 81.0 & 66.7 & 77.2 & 77.2 & 65.6 & 56.0 & 82.2 & 72.0 & 57.8 & 83.4 & 71.3 \\
    
    SHOT$_{\text{R50}}$ \cite{liang2020we} & 57.1 & 78.1 & 81.5 & 68.0 & 78.2 & 78.1 & 67.4 & 54.9 & 82.2 & 73.3 & 58.8 & 84.3 & 71.8 \\
    NRC$_{\text{R50}}$ \cite{yang2021exploiting} & 57.7 & \textbf{80.3} & 82.0 & 68.1 & \textbf{79.8} & 78.6 & 65.3 & 56.4 & 83.0 & 71.0 & 58.6 & \textbf{85.6} & 72.2 \\
    AAD$_{\text{R50}}$ \cite{yang2022attracting} & \textbf{59.3} & 79.3 & \underline{82.1} & 68.9 & \textbf{79.8} & \underline{79.5} & 67.2 & \underline{57.4} & 83.1 & 72.1 & 58.5 & \underline{85.4} & \underline{72.7} \\
    ELR(NRC)$_{\text{R50}}$~\cite{yi2023source} & 58.4 & 78.7 & 81.5 & \underline{69.2} & \underline{79.5} & 79.3 & 66.3 & \textbf{58.0} & 82.6 & \underline{73.4} & \underline{59.8} & 85.1 & 72.6 \\ 
    DIPE$_{\text{R50}}$~\cite{wang2022exploring} & 56.5 & 79.2 & 80.7 & \textbf{70.1} & \textbf{79.8} & 78.8 & 67.9 & 55.1 & \underline{83.5} & \textbf{74.1} & 59.3 & 84.8 & 72.5 \\
    A$^2$Net$_{\text{R50}}$~\cite{xia2021adaptive} & 58.4 & 79.0 & \textbf{82.4} & 67.5 & 79.3 & 78.9 & \underline{68.0} & 56.2 & 82.9 & \textbf{74.1} & \textbf{60.5} & 85.0 & \textbf{72.8} \\
    
    \hdashline\noalign{\vskip 0.5ex}
    \textbf{TAB}$_{\text{R50}}$ & \underline{58.9} & \underline{79.6} & 81.5 & 68.6 & 78.0 & \textbf{79.8} & \textbf{69.3} & 56.8 & \textbf{83.7} & 73.2 & 59.5 & 84.7 & \textbf{72.8} \\
    
    \hline
    ViT-L \cite{dosovitskiy2020image} & 75.3 & 88.5 & 91.4 & 85.3 & 89.7 & 89.9 & \underline{83.3} & 75.0 & 91.5 & 86.8 & 74.7 & \underline{92.0} & 85.3 \\
    SHOT$_{\text{ViT}}$ \cite{liang2020we} & \underline{80.9} & \underline{92.0} & \underline{91.9} & \textbf{89.9} & \underline{92.2} & 76.1 & 77.0 & \textbf{81.9} & \textbf{92.1} & \underline{88.9} & \textbf{82.8} & \textbf{93.9} & \underline{86.6} \\
    NRC$_{\text{ViT}}$ \cite{yang2021exploiting} & 79.7 & 91.0 & 91.8 & 85.8 & 89.7 & 91.7 & 42.7 & \underline{76.3} & 91.3 & 85.6 & \underline{82.6} & 88.3 & 83.0 \\
    AAD$_{\text{ViT}}$ \cite{yang2022attracting} & 66.3 & 91.2 & 91.7 & 89.7 & 90.5 & \underline{92.2} & 78.3 & 75.6 & 91.4 & 86.4 & 77.1 & \textbf{93.9} & 85.4 \\
    \hdashline\noalign{\vskip 0.5ex}
    \textbf{TAB}$_{\text{ViT}}$ & \textbf{81.3} & \textbf{92.7} & \textbf{93.2} & \underline{89.8} & \textbf{92.9} & \textbf{93.4} & \textbf{86.9} & 76.1 & \underline{91.7} & \textbf{89.4} & 80.9 & 89.7 & \textbf{88.2} \\
    \bottomrule
  \end{tabular}
  \end{adjustbox}
    \caption{Performance comparison of various SOTA methods on the \textit{Office-Home} dataset using both ResNet50 and ViT-Large backbones. Each column represents an experiment SRC$\veryshortarrow$TGT, while the rightmost column provides the average accuracy. The top results are highlighted in \textbf{bold}, while the runners-up are \underline{underlined}. All ViT-L outcomes were independently obtained by us.}
  \label{tab:officehome}
  \vspace{-0.2cm}
\end{table*}

\setcellgapes{-60pt}

\begin{table}[h!]

\centering
\makegapedcells

\begin{adjustbox}{width = 1\linewidth, center}

\setlength{\tabcolsep}{4pt}
\setlength{\extrarowheight}{1pt}
\setlength{\aboverulesep}{2pt}

\renewcommand\cellset{\renewcommand\arraystretch{0.3}%
                       \setlength\extrarowheight{4pt}}

\begin{tabular}{lccccccc}
\toprule
\textbf{Method} & P$\veryshortarrow$R &  P$\veryshortarrow$S & R$\veryshortarrow$P &  R$\veryshortarrow$S & S$\veryshortarrow$P &  S$\veryshortarrow$R & \textbf{Avg} \\
\midrule
ResNet50 \cite{he2016deep} & \mymakecellh{67.0}{0.6} & \mymakecellh{35.0}{1.0} & \mymakecellh{87.6}{0.1} & \mymakecellh{30.4}{0.9} & \mymakecellh{13.4}{1.8} & \mymakecellh{2.7}{0.8} & \mymakecellh{39.3}{0.5} \\
SHOT$_{\text{R50}}$ \cite{liang2020we} & \mymakecellh{\underline{78.5}}{0.2} & \mymakecellh{58.8}{2.0} & \mymakecellh{91.9}{0.2} & \mymakecellh{\underline{57.4}}{1.6} & \mymakecellh{58.9}{2.0} & \mymakecellh{\underline{44.6}}{2.9} & \mymakecellh{\underline{65.0}}{0.9} \\
AAD$_{\text{R50}}$ \cite{yang2022attracting} & \mymakecellh{76.7}{0.7} & \mymakecellh{53.5}{3.5} & \mymakecellh{\underline{92.1}}{0.2} & \mymakecellh{48.5}{3.3} & \mymakecellh{53.2}{3.0} & \mymakecellh{35.1}{3.2} & \mymakecellh{59.9}{1.5} \\
NRC$_{\text{R50}}$ \cite{yang2021exploiting} & \mymakecellh{77.2}{0.2} & \mymakecellh{\underline{60.8}}{0.7} & \mymakecellh{88.7}{0.3} & \mymakecellh{55.0}{1.9} & \mymakecellh{\underline{63.8}}{2.2} & \mymakecellh{44.0}{1.4} & \mymakecellh{64.9}{0.6} \\

\hdashline

\textbf{TAB}$_{\text{R50}}$ & \mymakecellh{\textbf{79.9}}{0.4} & \mymakecellh{\textbf{65.2}}{2.5} & \mymakecellh{\textbf{92.2}}{0.2} & \mymakecellh{\textbf{64.1}}{1.2} & \mymakecellh{\textbf{72.3}}{0.9} & \mymakecellh{\textbf{57.7}}{1.7} & \mymakecellh{\textbf{71.9}}{0.3} \\
\midrule
ViT-L \cite{dosovitskiy2020image} & \mymakecellh{93.5}{0.2} & \mymakecellh{68.9}{0.4} & \mymakecellh{97.4}{0.1} & \mymakecellh{66.2}{0.7} & \mymakecellh{93.2}{0.2} & \mymakecellh{87.2}{0.5} & \mymakecellh{84.4}{0.1} \\

SHOT$_{\text{ViT}}$ \cite{liang2020we} & \mymakecellh{\underline{94.5}}{0.3} & \mymakecellh{\textbf{87.6}}{0.7} & \mymakecellh{96.7}{2.6} & \mymakecellh{\textbf{86.9}}{0.4} & \mymakecellh{77.6}{42.9} & \mymakecellh{\textbf{91.7}}{1.8} & \mymakecellh{\underline{89.2}}{6.8} \\

AAD$_{\text{ViT}}$ \cite{yang2022attracting} & \mymakecellh{23.4}{26.6} & \mymakecellh{41.7}{23.4} & \mymakecellh{76.7}{42.3} & \mymakecellh{47.1}{4.9} & \mymakecellh{\underline{94.0}}{1.6} & \mymakecellh{12.4}{22.4} & \mymakecellh{49.2}{6.1} \\

NRC$_{\text{ViT}}$ \cite{yang2021exploiting}& \mymakecellh{93.2}{0.4} & \mymakecellh{83.3}{1.0} & \mymakecellh{\underline{97.5}}{0.2} & \mymakecellh{65.7}{35.9} & \mymakecellh{77.3}{42.4} & \mymakecellh{\underline{90.5}}{2.1} & \mymakecellh{84.6}{8.0} \\

\hdashline

\textbf{TAB}$_{\text{ViT}}$ & \mymakecellh{\textbf{94.6}}{0.3} & \mymakecellh{\underline{86.2}}{0.5} & \mymakecellh{\textbf{97.6}}{0.1} & \mymakecellh{\underline{86.0}}{1.2} & \mymakecellh{\textbf{96.5}}{0.5} & \mymakecellh{89.1}{1.3} & \mymakecellh{\textbf{91.7}}{0.3} \\
\bottomrule
\end{tabular}

\end{adjustbox}
   \caption{Multi-run (5 seeds) performance comparison of various SOTA methods on \textit{Adaptiope} dataset using both ResNet50 and ViT-Large backbones. The top results are highlighted in \textbf{bold}, while the runners-up are \underline{underlined}. All results have been obtained by us both for ResNet50 and ViT-L. \textbf{Note}: high standard deviations are due to the failure of methods for one or more seeds in the considered experiment.}
  \label{tab:adaptiope}
  \vspace{-0.2cm}
\end{table}

\indent\textbf{Adaptiope.} Table~\ref{tab:adaptiope} shows the results for this challenging benchmark, including both mean and standard deviation over five runs. On the ResNet50 architecture, our method significantly outperforms other techniques with a notable margin of $+6.9\%$. Furthermore, when utilizing the ViT-L architecture, our method continues to lead, registering an average accuracy of 91.7\%.

\indent\textbf{ImageCLEF-DA.} The results are provided in Table~\ref{tab:imageclef} (mean and std). On this small dataset, our method's performance is consistent with other techniques when implemented on both ResNet50 and ViT architectures. However, it is marginally outpaced by the NRC method, which achieves an average lead of $+0.4\%$.


\begin{table*}[h!]
\centering
\setlength{\tabcolsep}{2pt}
\setlength{\extrarowheight}{4pt}
\setlength{\aboverulesep}{2pt}
\setlength{\belowrulesep}{0pt}

\renewcommand\cellset{\renewcommand\arraystretch{0.3}%
                       \setlength\extrarowheight{4pt}}

\begin{adjustbox}{width = 0.99\linewidth, center}
\begin{tabular}{lccccccccccccr}
\toprule
    \textbf{Method} & B$\veryshortarrow$C &  B$\veryshortarrow$I & B$\veryshortarrow$P &  C$\veryshortarrow$B &  C$\veryshortarrow$I & C$\veryshortarrow$P &  I$\veryshortarrow$B &  I$\veryshortarrow$C & I$\veryshortarrow$P &  P$\veryshortarrow$B & P$\veryshortarrow$C & P$\veryshortarrow$I & \textbf{Avg} \\
    \midrule
    
ResNet50 \cite{he2016deep} & \mymakecell{90.0}{3.1} & \mymakecell{84.6}{2.8} & \mymakecell{68.0}{2.5} & \mymakecell{59.3}{1.3} & \mymakecell{83.9}{1.3} & \mymakecell{69.1}{1.8} & \mymakecell{60.6}{0.2} & \mymakecell{92.8}{0.8} & \mymakecell{75.2}{0.5} & \mymakecell{58.7}{1.4} & \mymakecell{91.1}{1.0} & \mymakecell{88.5}{2.2} & \mymakecell{76.8}{0.3} \\

SHOT$_{\text{R50}}$ \cite{liang2020we} & \mymakecell{\underline{96.6}}{0.6} & \mymakecell{\underline{92.8}}{0.6} & \mymakecell{77.7}{2.2} & \mymakecell{65.1}{0.5} & \mymakecell{\underline{92.8}}{0.3} & \mymakecell{78.0}{0.6} & \mymakecell{64.4}{0.8} & \mymakecell{96.5}{0.6} & \mymakecell{78.2}{0.6} & \mymakecell{64.4}{0.6} & \mymakecell{96.1}{0.4} & \mymakecell{92.5}{1.0} & \mymakecell{82.9}{0.1} \\

AAD$_{\text{R50}}$ \cite{yang2022attracting} & \mymakecell{96.5}{0.7} & \mymakecell{92.7}{0.6} & \mymakecell{\underline{78.0}}{1.2} & \mymakecell{\underline{65.7}}{0.7} & \mymakecell{92.6}{0.5} & \mymakecell{77.9}{0.7} & \mymakecell{\textbf{65.6}}{0.6} & \mymakecell{\underline{96.6}}{0.8} & \mymakecell{\underline{78.9}}{1.4} & \mymakecell{\underline{64.7}}{1.3} & \mymakecell{\underline{96.2}}{0.7} & \mymakecell{\textbf{93.2}}{0.7} & \mymakecell{\underline{83.2}}{0.2} \\

NRC$_{\text{R50}}$ \cite{yang2021exploiting} & \mymakecell{\underline{96.6}}{0.7} & \mymakecell{\textbf{93.2}}{0.3} & \mymakecell{\textbf{78.3}}{1.5} & \mymakecell{\textbf{65.9}}{0.8} & \mymakecell{\textbf{93.1}}{0.5} & \mymakecell{\underline{78.4}}{0.7} & \mymakecell{\underline{64.8}}{0.7} & \mymakecell{96.4}{0.5} & \mymakecell{\textbf{79.0}}{0.8} & \mymakecell{\textbf{65.2}}{1.0} & \mymakecell{96.1}{0.7} & \mymakecell{\underline{93.0}}{0.7} & \mymakecell{\textbf{83.3}}{0.0} \\

\hdashline

\textbf{TAB}$_{\text{R50}}$ & \mymakecell{\textbf{96.8}}{0.6} & \mymakecell{92.5}{0.0} & \mymakecell{77.8}{0.9} & \mymakecell{65.5}{0.9} & \mymakecell{92.2}{0.4} & \mymakecell{\textbf{78.5}}{0.3} & \mymakecell{64.5}{1.2} & \mymakecell{\textbf{97.0}}{0.2} & \mymakecell{78.2}{0.3} & \mymakecell{63.6}{0.6} & \mymakecell{\textbf{96.5}}{0.2} & \mymakecell{92.0}{0.1} & \mymakecell{82.9}{0.3} \\

\hline
ViT-L \cite{dosovitskiy2020image} & \mymakecell{96.2}{0.9} & \mymakecell{94.9}{0.7} & \mymakecell{78.2}{1.3} & \mymakecell{69.3}{1.1} & \mymakecell{94.6}{0.7} & \mymakecell{78.2}{1.2} & \mymakecell{69.9}{1.1} & \mymakecell{96.2}{0.4} & \mymakecell{81.2}{0.9} & \mymakecell{66.8}{0.9} & \mymakecell{92.6}{2.9} & \mymakecell{97.3}{0.8} & \mymakecell{84.6}{0.4} \\
SHOT$_{\text{ViT}}$ \cite{liang2020we} & \mymakecell{\underline{98.3}}{0.2} & \mymakecell{\underline{97.9}}{0.2} & \mymakecell{82.5}{0.4} & \mymakecell{\underline{71.5}}{1.3} & \mymakecell{98.0}{0.2} & \mymakecell{82.7}{0.3} & \mymakecell{72.2}{0.9} & \mymakecell{98.2}{0.4} & \mymakecell{82.8}{0.5} & \mymakecell{72.5}{0.9} & \mymakecell{98.3}{0.3} & \mymakecell{98.3}{0.2} & \mymakecell{87.8}{0.3} \\
AAD$_{\text{ViT}}$ \cite{yang2022attracting} & \mymakecell{95.4}{6.9} & \mymakecell{\textbf{98.0}}{0.2} & \mymakecell{\underline{83.1}}{0.6} & \mymakecell{70.9}{3.5} & \mymakecell{98.0}{0.2} & \mymakecell{\underline{82.8}}{0.5} & \mymakecell{\underline{74.2}}{1.0} & \mymakecell{\underline{98.3}}{0.5} & \mymakecell{83.4}{0.6} & \mymakecell{\textbf{74.3}}{1.1} & \mymakecell{96.9}{3.5} & \mymakecell{98.3}{0.4} & \mymakecell{87.8}{1.1} \\
NRC$_{\text{ViT}}$ \cite{yang2021exploiting} & \mymakecell{\underline{98.3}}{0.2} & \mymakecell{\textbf{98.0}}{0.2} & \mymakecell{\textbf{83.2}}{0.4} & \mymakecell{\textbf{74.4}}{0.5} & \mymakecell{\textbf{98.4}}{0.3} & \mymakecell{82.6}{0.7} & \mymakecell{\textbf{74.3}}{0.6} & \mymakecell{98.2}{0.5} & \mymakecell{\textbf{83.6}}{0.3} & \mymakecell{\underline{73.5}}{0.8} & \mymakecell{\underline{98.4}}{0.5} & \mymakecell{\textbf{98.5}}{0.4} & \mymakecell{\textbf{88.4}}{0.2} \\

\hdashline

\textbf{TAB}$_{\text{ViT}}$ & 
\mymakecell{\textbf{98.8}}{0.1} & \mymakecell{\textbf{98.0}}{0.3} & \mymakecell{82.9}{0.4} & \mymakecell{70.6}{3.1} & \mymakecell{$\underline{98.3}$}{0.2} & 
\mymakecell{$\textbf{82.9}$}{0.1} & \mymakecell{$72.7$}{0.2} & \mymakecell{\textbf{98.8}}{0.2} & \mymakecell{\underline{83.4}}{0.5} & \mymakecell{71.8}{0.8} & \mymakecell{\textbf{98.7}}{0.2} & \mymakecell{\underline{98.4}}{0.2} & \mymakecell{\underline{88.0}}{0.3} \\

\bottomrule

\end{tabular}
\end{adjustbox}

  \caption{Multi-run (5 seeds) performance comparison of various SOTA methods on \textit{ImageCLEF-DA} dataset using both ResNet50 and ViT-Large backbones. The top results are highlighted in \textbf{bold}, while the runners-up are \underline{underlined}. All results have been obtained by us both for ResNet50 and ViT-L.}
  \label{tab:imageclef}
   \vspace{-0.2cm}
\end{table*}


\subsection{Analysis}\label{sub:analysis}
The results highlight that our method, with its inherently simple yet effective pseudo-labeling approach and clear design, achieves performance that is on par with or even surpasses state-of-the-art methods across the examined benchmarks. Significantly, our approach outperforms competitors in 3 out of 4 datasets when using the ViT-Large architecture. It is especially notable that our method exceeds others substantially in the challenging Adaptiope benchmark when employing ResNet50, showcasing its robustness. 


%
%
%


\indent\textbf{Ablation Study.} \cref{tab:ablation} demonstrates the effectiveness of our pseudo-labeling procedure, FTSP, which achieves strong performance on Office31 and Office-Home datasets. The addition of pseudo-label refinement phase (PR) and our TSAL objective further improve this performance. \cref{fig:temperature_scaling} presents the impact of TSAL's temperature scaling on both the discriminability and diversity terms during the A $\to$ W experiment of Office31. As expected, as training progresses, the discriminability term rises due to the increasing temperature. This counteracts the network's over-confidence and keeps the pseudo-label information relevant. In contrast, without temperature scaling, the diversity term drops close to its lowest possible value in the first epoch ($\log(1/31) \simeq -3.43$, where 31 is the number of classes). But with temperature scaling, the network's predictions are sharpen, allowing for a higher diversity value. These effects result in an improved adaptation (from 92.8\%, without temperature scaling, to 94.7\% accuracy in the considered experiment). A comprehensive ablation study involving different classifiers for FTSP, hyperparameters and values of trusted samples (K) is presented in the supplementary material. \\
\indent\textbf{Robustness.} We evaluated our algorithm across four different benchmarks. For ImageCLEF-DA (Tab.~\ref{tab:imageclef}) and Adaptiope (Tab.~\ref{tab:adaptiope}), experiments were conducted with 5 seeds each. On the small ImageCLEF-DA, all methods we considered show stable results. However, in Adaptiope with its pronounced domain gaps, certain methods encountered difficulties. AAD did not achieve satisfactory results for both ResNet50 and ViT-L architectures. NRC yielded less than optimal results for both, and while SHOT performed well with ViT-L, it faced challenges with ResNet50. In contrast, our proposed method demonstrated consistent performance across both architectures, surpassing other approaches. \\
\indent \textbf{Remarks and limitations.} The experiments on all datasets were conducted using constant, and potentially not-optimal, hyperparameters. A \textbf{post hoc} ablation study (available in the supplementary material) demonstrates that TAB is more robust to hyperparameters variations than other methodologies. Moreover, it suggests also that by choosing different configurations, performance can be further enhanced. For example, with ResNet50, TAB achieves an average accuracy of $\mathbf{90.3\%}$ on Office31 and $\mathbf{72.9\%}$ on Office-Home using $K = 5$. While our approach already shows competitive and superior results when compared with SOTA methods, we recognize that adopting unsupervised hyperparameter selection techniques (e.g., \cite{saito2021tune}), as utilized by other approaches \cite{yang2022attracting, yang2021exploiting}, could further boost our method's performance. We leave this exploration for future work.

\noindent 
\begin{minipage}[t]{0.44\textwidth} 
\vspace{-2.80cm}
\centering

\centering
\setlength{\tabcolsep}{2pt}
\setlength{\extrarowheight}{-100pt}
\begin{tabular}{ccccc}
\toprule
FSPL & PR & TSAL & O.31 & O.Home \\
\midrule
\checkmark & & & 89.4 & 72.0 \\
\checkmark & \checkmark & & 89.6 & 72.2 \\
\checkmark & & \checkmark & 89.6 & 72.4 \\
\checkmark & \checkmark & \checkmark & \textbf{89.9} & \textbf{72.8} \\
\bottomrule
\end{tabular}
\vspace{0.52cm}
\captionof{table}{The introduction of Pseudo-label Refinement (PR), i.e. Label Deletion + Label Spreading, enhance the performance. The addition of TSAL give an additional boost.} 
\label{tab:ablation}
\end{minipage}%
\hfill
\begin{minipage}[t]{0.5\textwidth} 
\centering
\includegraphics[width=0.95\linewidth]{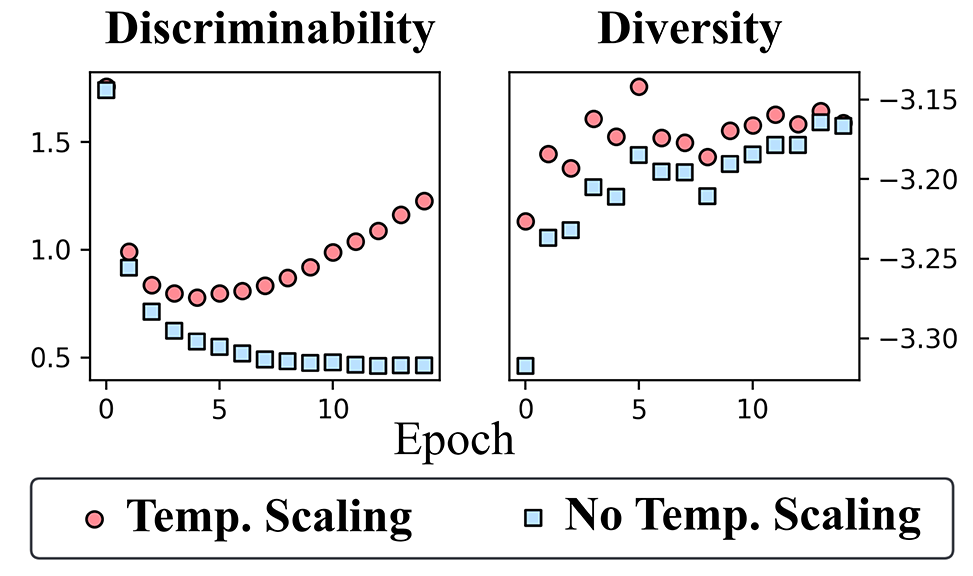}
\captionof{figure}{Discriminability and Diversity terms of the SF-UDA objective averaged across each training epoch with and without temperature scaling. The plot shows the experiment Amazon $\to$ Webcam of Office31.} 
\label{fig:temperature_scaling}
\end{minipage}

\vspace{-0.4cm}
\section{Conclusion}
\label{sec:conclusion}

In this work, we introduced \textbf{Trust And Balance} (TAB), a novel, simple and effective method for SF-UDA in image classification. It is characterized by two key innovations: Few Trusted Samples Pseudo-labeling, which computes high quality pseudo-labels for the target domain, and Temperature Scaled Adaptive Loss, which balances the diversity, the discriminability and the pseudo-labels significance in the objective.
The empirical evaluations show that TAB, despite its simple design, obtains performance similar to or better than SOTA methods in all benchmarks considered and an increased robustness. 

\section{Training Details}
\label{sec:train_details_full}

As outlined in the main text, we employed standard procedures and hyperparameters drawn from existing literature.

\vspace{0.3cm}

\indent\textbf{Pre-training.} For the ResNet50 architecture, we utilized weights pre-trained on ImageNet as provided by TorchVision \cite{torchvision2016}. For the ViT-Large architecture, weights pre-trained on ImageNet21k from the timm library \cite{rw2019timm} were used.

\vspace{0.3cm}

\indent\textbf{Input pipeline and augmentations.} Initially, images are resized to dimensions $256 \times 256$. For training purposes, a random crop yielding a dimension of $224 \times 224$ is executed along with the application of a random horizontal flip. During evaluation, a centered crop is applied. Subsequent to these transformations, images undergo normalization based on the mean and standard deviation values from ImageNet, consistent with the pre-training setup of the backbones.

\vspace{0.3cm}

\indent\textbf{Source fine-tuning.} The fine-tuning on the source domain employs SGD with Nesterov Momentum of $0.9$ and an $L2$ penalty of $10^{-3}$. A batch size of 64 is used, with learning rates initialized to $10^{-3}$ for the pre-trained backbone and $10^{-2}$ for the additional bottleneck and classifier with freshly initialized weights. We employ a cross-entropy objective with label smoothing \cite{muller2019does} using a factor of $0.1$ and clip gradient norms at $5.0$. The learning rate adjustment follows the schedule:

\begin{equation}
    lr(t) = lr(0) * \left( 1 + 10 \cdot \frac{t}{T} \right)^{-0.75}
\end{equation}

where $lr(0)$ represents the initial learning rate, $T$ the total number of training steps and $t$ the current training step. The dataset is partitioned into training (85\%) and validation (15\%) subsets. Training continues for 100 epochs, and after each epoch, validation accuracy is assessed. Model weights achieving the highest validation accuracy are retained. Distributed fine-tuning is executed on 4 Nvidia V100 16GB GPUs.

\vspace{0.3cm}

\indent\textbf{Target adaptation.} This phase uses a batch size of 64 with similar learning rates and scheduling as source fine-tuning for 15 epochs, but the classifier $\gamma$ remains fixed (lr set to 0). At every epoch, FTSP computes pseudo-labels, and by default, Multinomial Logistic Regression serves as the classifier. We employ the MLR classifier from Scikit-Learn \cite{scikit-learn} with default settings and minimal strength $L2$ regularization ($C = \frac{1}{\lambda} = 1000$). We subsequently apply our Pseudo-label Refinement (PR): we remove the 20\% least confident pseudo-labels from each class based on predicted probabilities via the \textbf{label deletion} step. In the \textbf{label completion} phase, Label Spreading \cite{zhou2003learning} (from Scikit-Learn) with default hyperparameters (RBF kernel with $gamma=20$) is used. The TSAL objective, described in the main text, is then minimized. Additionally, MixUp \cite{zhang2017mixup} regularization is utilized, where the mixing ratio is sourced from a Beta distribution with parameters $\alpha = \beta = 0.3$. Since no label is available in this training phase the adapted model is the one obtained after the whole 15 epochs training that is directly tested with labels. Target adaptation takes place on a singular GPU: NVIDIA V100 16GB for ResNet50 and NVIDIA A100 80GB for ViT-Large.

\section{Few Trusted Samples Pseudo-labeling: an ablation study}
\label{sec:app_ablation}

In this section, we present an ablation study centered around FTSP, in particular we analyze main hyperparmater introduced by our algorithm: the number of trusted samples $K$. 

This study will show that our methodology is very robust to this hyperparameter in the range considered with a accuracy change of only $0.7\%$ for Office-31 and $0.3\%$ for Office-Home ($0.7\%$ without Pseudo-Label refinement). 
To give a comparison, in our experiments, changing the $K$ and $KK$ hyperparamters of NRC~\cite{yang2021exploiting} in the range presented in the original paper can result in a potential accuracy degradation of $2.0\%$ for Office-31 and $1.3\%$ for Office-Home. Similarly, for AAD~\cite{yang2022attracting} changing $\beta$ and $K$ hyperparameters can result in a accuracy change of $3.0\%$ for Office-31 and $6.8\%$ for Office-Home.

Additionally we present an analysis on the contribution of the Pseudo-label Refinement phase and on different types of classifiers in the FTSP procedure.

\noindent\textbf{Remark.} It is crucial to clarify that the hyperparameters used for the experiments presented in the main text were kept fixed. The ablation experiments described herein have been conducted subsequently to those in the main text to prevent potential evaluation biases. 

\vspace{0.3cm}

\indent \textbf{Number of Trusted Samples and Pseudo-label Refinement.} Throughout the experiments, we consistently set the number \( K \) of trusted samples (per class) to 3 for ResNet50 and 7 for ViT. We selected \( K = 3 \) for ResNet50 aiming for a constrained set of trusted samples, ensuring these samples had accurate labels for the classifier training in FTSP. This approach stems from the understanding that a limited number of samples can adequately train a classifier, which then generalizes effectively across the entire target domain, as our empirical results confirm. Given that (according to the ImageNet benchmark) ViT-Large is a better performing model compared to ResNet50, it also exhibits superior out-of-distribution generalization as shown in \cite{maracani2023key}. Consequently, we followed this intuition and we increased \( K \) to 7 for ViT-L, anticipating enhanced predictions and greater model reliability.
In Table~\ref{tab:app_ablation3}, we provide an ablation study focusing on the value of \( K \) for ResNet50 and on the effects of the Pseudo-labeling Refinement phase that we propose. As it is possible to observe in the table both for Office31 and Office-Home the algorithm is robust to the choice of $K$ in the range considered and the Pseudo-Labeling refinement provides in almost all cases benefits in terms of accuracy. Additionally the value of $K=5$ seems to be the best choice for these datasets considering our classifier (MLR), increasing marginally the results reported in the main text (with $K=3$). For completeness we report in Table~\ref{tab:office31_app_abl} and \ref{tab:officehome_app_abl} the comparison of the best performing SOTA methods, with our proposed method with optimal $K$. 

\vspace{0.3cm}

\indent \textbf{The choice of the classifier.} We present an ablation study about the choice of the classifier for FTSP. In particular, in this study, we focused on the FTSP methodology \textbf{without Pseudo-label Refinement} and we optimized the TSAL objective as stated in the main text. We evaluated the following classifiers: Support Vector Machine Classifier with RBF (SVC-R) and Linear (SVC-L) kernels, Multinomial Logistic Regression (MLR), and Linear Discriminant Analysis (LDA). For these classifiers, we examined a variety of L2 regularization strengths, regulated by the C hyperparameter in Scikit-Learn (where the value of C is inversely proportional to the strength of regularization). We also assessed different shrinkage values for LDA.

Results for the Office-Home dataset are provided in table \ref{tab:app_ablation}, and the outcomes for the Office31 dataset are presented in table \ref{tab:app_ablation2}. For both datasets, it is evident that MLR, particularly with weaker regularization, outperforms SVC. Furthermore, the Linear Discriminant Analysis Classifier surpasses MLR in performance for the datasets examined.

\begin{table}[h!]
\centering
\setlength{\tabcolsep}{2pt}
\begin{tabular}{lcccccc}
\toprule
\textbf{Classifier} & K=2 & K=3 & K=5 & K=7 & K=10 \\
\midrule
Office31 (w/o PR) & 89.0 & 89.6 & \textbf{89.7} & 89.2 & 88.9 \\
Office-Home (w/o PR) & 72.3 & 72.4 & \textbf{72.9} & 72.5 & 72.3 \\
\midrule
Office31 (with PR) & 89.6 & 89.9 & \textbf{90.3} & 89.9 & 89.8 \\
Office-Home (with PR) & 72.6 & 72.8 & \textbf{72.9} & 72.8 & 72.5 \\
\bottomrule
    
\end{tabular}
\caption{\textbf{Ablation on K with and without Pseudo-label Refinement}: Average accuracy of FTSP+TSAL using MLR with C=1000 for Office31 and Office-Home using different values of K.}
  \label{tab:app_ablation3}
\end{table}

\begin{table}[h!]
  \centering
  \setlength{\tabcolsep}{2pt}


\begin{tabular}{lccccccc}
    \toprule
    \textbf{Method} & A$\veryshortarrow$D &  A$\veryshortarrow$W & D$\veryshortarrow$A &  D$\veryshortarrow$W & W$\veryshortarrow$A &  W$\veryshortarrow$D & \textbf{Avg} \\
    \midrule
    DIPE$_{\text{R50}}$~\cite{wang2022exploring} & \textbf{96.6} & 93.1 & 75.5 & 98.4 & \underline{77.2} & 99.6 & \underline{90.1} \\
    A$^2$Net$_{\text{R50}}$~\cite{xia2021adaptive} & 94.5 & 94.0 & 76.7 & \textbf{99.2} & 76.1 & \textbf{100.0} & \underline{90.1} \\
    \hdashline\noalign{\vskip 0.5ex}
    \textbf{TAB}$_{K=3}$ & 94.4 & \textbf{94.7} & \underline{76.9} & 97.4 & 76.0 & \underline{99.8} & 89.9 \\
    \textbf{TAB}$_{K=5}$ & \underline{94.6} & \underline{94.1} & \textbf{77.5} & \underline{98.7} & \textbf{77.3} & 99.4 & \textbf{90.3} \\
    \bottomrule
\end{tabular}

  \caption{Comparison of TAB (with $K = 3$ and $K = 5$) with two state-of-the-art methods on the \textbf{Office31} dataset using ResNet50. The top results are highlighted in \textbf{bold}, while the runners-up are \underline{underlined}.}
  \label{tab:office31_app_abl}
\end{table}

\begin{table}[t!]
\centering
\setlength{\tabcolsep}{2pt}
\begin{tabular}{cccccc}
\toprule
\textbf{Classifier} & C=0.1 & C=1.0 & C=1000 & S=0.5 & S=0.99 \\
\midrule
SVC-R & 68.9 & 71.5 & 71.3 & --- & --- \\
SVC-L & 70.0 & 72.0 & 71.7 & --- & --- \\
MLR  & 72.3 & 72.3 & 72.4 & --- & --- \\
LDA & --- & --- & --- & 72.6 & \textbf{72.7} \\ 
\bottomrule
   
\end{tabular}
\caption{\textbf{Ablation on Office-Home}: Evaluation of SVM Classifiers using RBF (SVC-R) and Linear (SVC-L) kernels, and Multinomial Logistic Regression (MLR) across varying L2 regularization strengths. Here, the C hyperparameter is inversely proportional to regularization strength. Linear Discriminant Analysis with different shrinkage values (S) is also assessed. The value of trusted samples $K$ is set to $3$.}

  \label{tab:app_ablation}
\end{table}

\begin{table}[h!]
\centering
\setlength{\tabcolsep}{2pt}
\begin{tabular}{cccccc}
\toprule
\textbf{Classifier} & C=0.1 & C=1.0 & C=1000 & S=0.5 & S=0.99 \\
\midrule
SVC-R & 87.4 & 89.2 & 89.4 & --- & --- \\
SVC-L & 88.2 & 89.1 & 89.3 & --- & ---\\
MLR  & 89.0 & 89.2 & 89.6 & --- & --- \\
LDA & --- & --- & --- & 89.2 & \textbf{89.8} \\ 
\bottomrule
   
\end{tabular}
\caption{\textbf{Ablation on Office31}: Evaluation of SVM Classifiers using RBF (SVC-R) and Linear (SVC-L) kernels, and Multinomial Logistic Regression (MLR) across varying L2 regularization strengths. Here, the C hyperparameter is inversely proportional to regularization strength. Linear Discriminant Analysis with different shrinkage values (S) is also assessed.The value of trusted samples $K$ is set to $3$.}
  \label{tab:app_ablation2}
\end{table}

\begin{table*}[h!]
  \centering
  \setlength{\tabcolsep}{4pt}

  \begin{adjustbox}{width = 0.99\textwidth, center}
  \begin{tabular}{lccccccccccccc}
    \toprule
    \textbf{Method} & A$\veryshortarrow$C &  A$\veryshortarrow$P & A$\veryshortarrow$R &  C$\veryshortarrow$A &  C$\veryshortarrow$P & C$\veryshortarrow$R &  P$\veryshortarrow$A &  P$\veryshortarrow$C & P$\veryshortarrow$R &  R$\veryshortarrow$A & R$\veryshortarrow$C & R$\veryshortarrow$P & \textbf{Avg} \\
    \midrule
    AAD$_{\text{R50}}$ \cite{yang2022attracting} & \textbf{59.3} & 79.3 & \underline{82.1} & \textbf{68.9} & \underline{79.8} & \underline{79.5} & 67.2 & \underline{57.4} & \underline{83.1} & 72.1 & 58.5 & \textbf{85.4} & 72.7 \\
    A$^2$Net$_{\text{R50}}$~\cite{xia2021adaptive} & 58.4 & 79.0 & \textbf{82.4} & 67.5 & 79.3 & 78.9 & 68.0 & 56.2 & 82.9 & \textbf{74.1} & \textbf{60.5} & 85.0 & \underline{72.8} \\
    \hdashline\noalign{\vskip 0.5ex}
    \textbf{TAB}$_{K=3}$ & \underline{58.9} & \underline{79.6} & 81.5 & \underline{68.6} & 78.0 & \textbf{79.8} & \underline{69.3} & 56.8 & \textbf{83.7} & \underline{73.2} & 59.5 & 84.7 & \underline{72.8} \\
    \textbf{TAB}$_{K=5}$ & 58.3 & \textbf{80.3} & 81.5 & 67.3 & \textbf{81.0} & 78.4 & \textbf{69.8} & \textbf{57.8} & 83.0 & 72.0 & \underline{60.1} & \underline{85.3} & \textbf{72.9} \\
    \bottomrule
  \end{tabular}
  \end{adjustbox}
  \caption{Comparison of TAB (with $K = 3$ and $K = 5$) with two state-of-the-art methods on the \textbf{Office-Home} dataset using ResNet50. The top results are highlighted in \textbf{bold}, while the runners-up are \underline{underlined}.}
  \label{tab:officehome_app_abl}
\end{table*}

\clearpage

\section{Efficiency of the Proposed Approach}
\label{tabsec:aap_efficiency}
The computational demands for executing the optimization of TSAL objective during the target adaptation stage are comparable to those encountered in standard supervised learning. These demands are influenced by several factors, including the chosen model architecture (backbone), the use of hardware accelerators, the software implementation, and the volume of data being processed. 

The additional computational steps introduced by TAB at each adaptation epoch are primarily for generating pseudo-labels, which involves the following processes:
\begin{enumerate}
\item Extraction of features and prediction probabilities for images from the target domain.
\item Selection of a Few-Trusted Samples dataset based on model predictions.
\item Training of a classifier on the Few-Trusted Samples dataset.
\item Use of the trained classifier to generate pseudo-labels for the target domain.
\item Optional application of Label Spreading to further refine the pseudo-labels.
\end{enumerate}

The majority of the computational effort and time is allocated to the first step, which is a common requirement across many SF-UDA algorithms. Steps 2 through 5, which are unique to the TAB approach, require comparatively minimal time relative to feature extraction. For context, in our experiments, steps 2 through 5, implemented using Scikit-learn algorithms running on CPU, took only a few seconds, whereas feature extraction could take minutes, especially with large architectures, extensive datasets, or in the absence of hardware acceleration. Our publicly available code includes functions that facilitate feature extraction (and model optimization) through distributed computing across multiple GPUs, significantly enhancing time-efficiency.

In summary, the computational times for our approach are on par with other state-of-the-art methodologies documented in the literature, such as SHOT~\cite{liang2020we}, AAD~\cite{yang2022attracting}, and NRC~\cite{yang2021exploiting} when models are adapted using a single hardware accelerator. Moreover, our distributed computing implementation further optimizes the performance of TAB, especially when additional GPUs are utilized, enabling experiments with larger models and datasets.

%
%




\bibliographystyle{splncs04}
\bibliography{main}
\end{document}